\documentclass[11pt,a4paper]{article}
\usepackage[hyperref]{emnlp2020}
\usepackage{times}
\usepackage{latexsym}

\usepackage{microtype}

\usepackage{graphicx}
\usepackage{amsmath}
\usepackage{amsthm}
\usepackage{algorithm}
\usepackage{algorithmic}
\usepackage{booktabs}

\usepackage{multirow}
\usepackage{color}
\usepackage{amsfonts}
\usepackage{bm}
\usepackage{CJKutf8}
\usepackage{array}
\usepackage{etoolbox}
\usepackage[bottom]{footmisc}
\newcolumntype{L}[1]{>{\raggedright\let\newline\\\arraybackslash\hspace{0pt}}m{#1}}
\newcolumntype{C}[1]{>{\centering\let\newline\\\arraybackslash\hspace{0pt}}m{#1}}
\newcolumntype{R}[1]{>{\raggedleft\let\newline\\\arraybackslash\hspace{0pt}}m{#1}}

\aclfinalcopy 
\def\aclpaperid{478} 

\title{Dialogue Distillation: Open-Domain Dialogue Augmentation Using Unpaired Data}

\author{Rongsheng Zhang$^{1}$\thanks{\quad Equal contribution. Order determined by swapping the one in \citet{zheng2019pre}}, Yinhe Zheng$^{2,3}$\footnotemark[1], Jianzhi Shao$^4$\thanks{\quad Work performed while at Fuxi AI Lab, NetEase Inc.}, Xiaoxi Mao$^1$, \\
\textbf{Yadong Xi$^1$, Minlie Huang$^2$\thanks{\quad Corresponding Author: aihuang@tsinghua.edu.cn}} \\
  $^1$ {\normalsize Fuxi AI Lab, NetEase Inc., Hangzhou, China} \\
  $^2$ {\normalsize Department of Computer Science and Technology, Institute for Artifical Intelligence, State Key} \\
  {\normalsize Lab of Intelligent Technology and Systems, Beijing National Research Center for} \\
  {\normalsize Information Science and Technology, Tsinghua University, Beijing, China.} \\
  $^3$ {\normalsize Samsung Research China - Beijing (SRC-B), Beijing, China} $^4$ {\normalsize Alibaba Group, Hangzhou, China} \\
  \texttt{\normalsize zhangrongsheng@corp.netease.com, yh.zheng@samsung.com}}
\date{}

\begin{document}
\maketitle
\begin{abstract}
Recent advances in open-domain dialogue systems rely on the success of neural models that are trained on large-scale data. However, collecting large-scale dialogue data is usually time-consuming and labor-intensive. To address this data dilemma, we propose a novel data augmentation method for training open-domain dialogue models by utilizing unpaired data. Specifically, a data-level distillation process is first proposed to construct augmented dialogues where both post and response are retrieved from the unpaired data. A ranking module is employed to filter out low-quality dialogues. Further, a model-level distillation process is employed to distill a teacher model trained on high-quality paired data to augmented dialogue pairs, thereby preventing dialogue models from being affected by the noise in the augmented data.
Automatic and manual evaluation indicates that our method can produce high-quality dialogue pairs with diverse contents, and the proposed data-level and model-level dialogue distillation can improve the performance of competitive baselines.
\end{abstract}

\begin{CJK}{UTF8}{gbsn}
\section{Introduction}
Open-domain dialogue systems have attracted much research attention ~\cite{shum2018,huang2019}, thanks to the success of neural generation models trained with large-scale data. Existing research has been endeavored to address various aspects in dialogue systems, such as modeling persona~\cite{qian2018assigning,zheng2019personalized,Zhang2018Personalizing},
expressing emotion~\cite{zhou2018emotional}, or generating knowledge-grounded dialogues~\cite{ghazvininejad2018knowledge,zhou2018commonsense,zhou-etal-2020-kdconv}.

In general, training neural open-domain dialogue models requires a large amount of high-quality \emph{paired data}, e.g., post-response pairs, which are usually labor-intensive and time consuming to collect. A feasible solution to this data dilemma is to use data augmentation techniques, which are popular in various research areas such as computer vision~\cite{cubuk2018autoaugment} or machine translation~\cite{sennrich2015improving}. Nevertheless, this technique is rarely investigated in the study of open-domain dialogues, and few existing approaches are specifically designed for either the generation-based dialogue models~\cite{li2019insufficient} or the retrieval-based dialogue models~\cite{du2018data}. Moreover, existing data augmentation approaches only take a set of paired data as input without considering to utilize unpaired data.

\begin{figure}[t]
    \centering
    \includegraphics[width=220px]{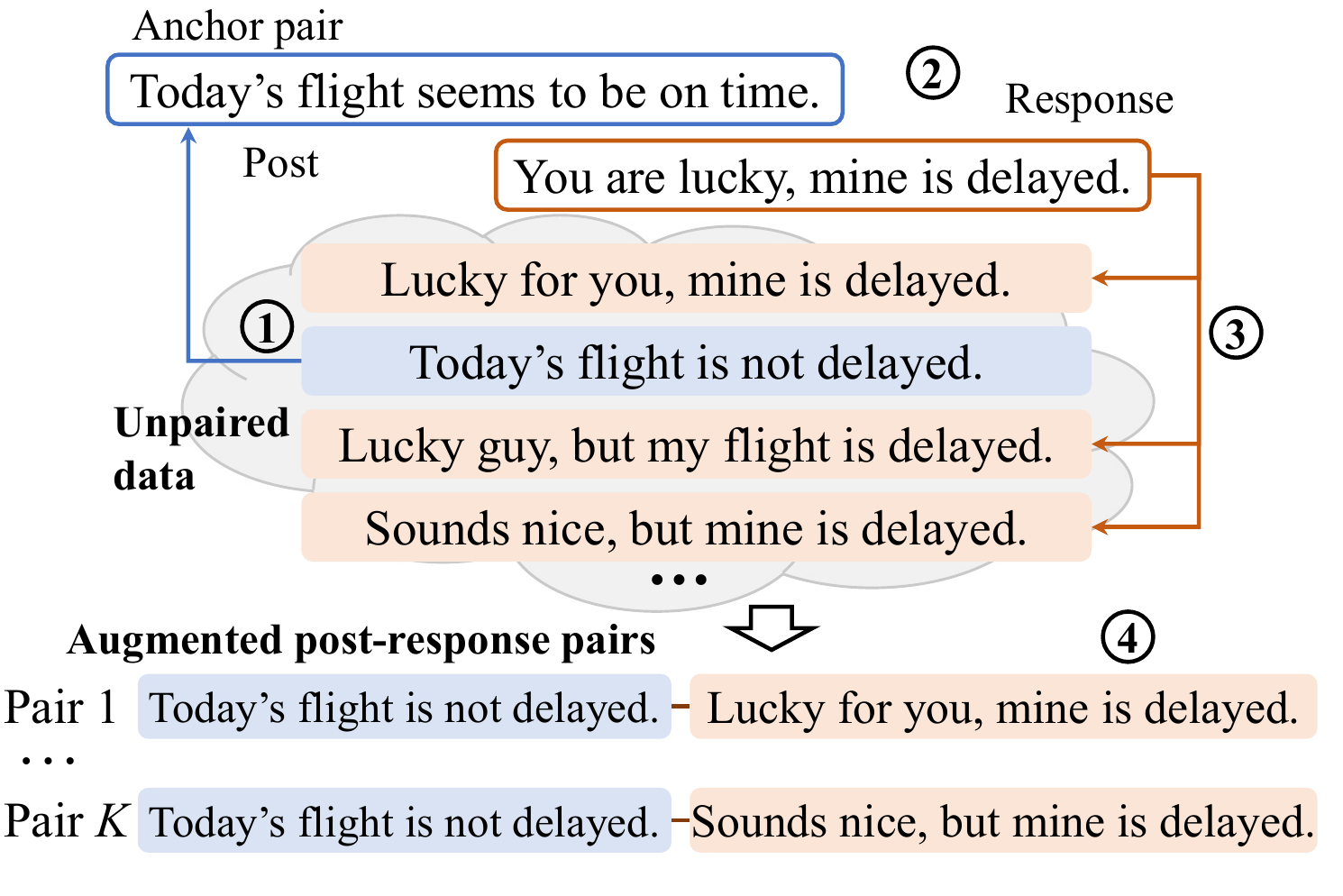}
    \caption{Process of constructing augmented post-response pairs. The sentence in blue rectangle is used to match the anchor pair and the corresponding response is then used to retrieve similar sentences in unpaired data. Each augmented pair contains two sentences both from unpaired data.}
    \label{fig:aug_with_unpaired}
\end{figure}

As a matter of fact, high-quality \emph{unpaired data}, i.e., non-conversational texts, are generally easier to collect compared to high-quality dialogue pairs. Specifically, these unpaired data provide us a rich bank of alternative expressions for different contents. It is thus feasible to augment the training dialogue pairs utilizing  sentences extracted from the unpaired data. As shown in Figure~\ref{fig:aug_with_unpaired}, we can extract various sentences from the unpaired data that are similar to a given post-response pair (i.e., anchor pair). Augmented pairs that carry richer expressions can be then constructed by combining these extracted sentences. To the best of our knowledge, there are no previous studies for open-domain dialogues that try to construct augmented dialogue pairs by utilizing retrieved unpaired data.

In this paper, we propose a novel data augmentation method ``\textbf{Dialogue Distillation}'' to improve the performance of open-domain dialogue models by utilizing unpaired data. Our method involves two phases of distillation. The first phase is at the data level as it constructs (i.e., distills) post-response pairs by matching sentences retrieved from a set of unpaired data. Specifically, given a set of training pairs $\{\langle x_i,y_i \rangle\}$, a randomly selected sentence $s$ is firstly used as a query to retrieve the most similar $x_i$, and then the corresponding $y_i$ are used as queries to retrieve similar $s_i$ from the unpaired data. Augmented pairs $\langle s,s_i \rangle$ are then constructed and filtered using a ranking module. Note that different from previous approaches, \textit{the post and response sentences that constitute an augmented pair are both from the unpaired data}, which are human written and thereby fluent and content-rich. The second phase is at the model-level as it distills a teacher model using the augmented data. Specifically, we borrow the idea of knowledge distillation~\cite{hinton2015distilling} to first train a teacher model on a set of high-quality dialogue pairs, and then distill the dialogue model by mimicking the distribution produced by the teacher model on the augmented data to prevent the final dialogue models from being affected by the noise in the augmented data.

Automatic and manual evaluation results indicate that our data-level distillation process can produce high-quality post-response pairs that are content-rich, and our model-level distillation process can better utilize these augmented data to improve the performance of both retrieval-based and generation-based open-domain dialogue models.

Our contributions are summarized as follows:

\textbf{1)} We propose a data-level and model-level distillation method for open-domain dialogue models. The data-level distillation constructs new post-response pairs where both post and response are retrieved from unpaired data, and the model-level distillation distills a teacher model trained on high quality paired data to augmented pairs. To the best of our knowledge, this is the first attempt to augment open-domain dialogue pairs by utilizing the retrieved unpaired data.

\textbf{2)} Automatic and manual evaluation shows that the augmented pairs produced by our method are content-rich, and these augmented data can be used to improve the performance of both generation-based and retrieval-based dialogue models.

\section{Related Work}

There are two major categories of open-domain dialogue models: 1) retrieval-based models, which retrieve the best matching response from the pre-collected dialogues~\cite{lu2013deep}; and 2) generation-based models, which decode responses from a learned distribution~\cite{sutskever2014sequence,vinyals2015neural}. Recent advances in these two categories all focus on DNN-based data-driven methods~\cite{huang2019}.

Data augmentation is an effective approach to boost the performance of neural models. It has been explored in various NLP tasks, such as text classification~\cite{wei2019eda,zheng2020ood}, machine reading comprehension~\cite{yu2018qanet} and machine translation~\cite{sennrich2015improving}. Although proved to be effective, this technique is rarely investigated in open-domain dialogue models. Few existing approaches are restricted to only take the dialogue pairs as their inputs~\cite{li2019insufficient,zhao2017generative,cai-etal-2020-data}, whereas unpaired texts, i.e., sentences without replies, are not utilized.

Note that the pre-training based methods~\cite{devlin2018bert,radford2019language,golovanov2019large,zheng2019pre} share a similar motivation with our study, i.e., to boost the performance of neural NLP models utilizing unlabeled (i.e., unpaired) texts. Nevertheless, the data augmentation method proposed in our study can be regarded as a supplement to these pre-training approaches. Experiments demonstrate that our method can be used to improve the performance of dialogue models even if these models are initialized with strong pre-trained models.

Our study is also related to the knowledge distillation method~\cite{hinton2015distilling}, which also employs a teacher model and tries to minimize the KL divergence between the teacher distribution and the model distribution. The most related work in this branch compared to ours was done by \citet{kim2016sequence}. However, their methods do not utilize unpaired data, and the augmented data are decoded from a probability model using beam search. Whereas our method tries to utilize the unpaired data, and the augmented data are generated by aligning human produced sentences.

There are also works that try to utilize retrieved non-conversational texts to improve the diversity of the dialogue model \cite{wu2019response,cai2019skeleton,zhu2019retrieval,su2020diversifying}. However, most of these studies focus on extracting templates from these non-conversational texts rather than generating augmented pairs, and they typically use specifically designed model structures.  Nevertheless, the data augmentation method proposed in our study can be used in combination with any dialogue models to improve the performance.

\section{Data-level Distillation}
\begin{figure}[t]
  \centering
  \setlength{\belowcaptionskip}{-0.1cm}
  \includegraphics[width=210px]{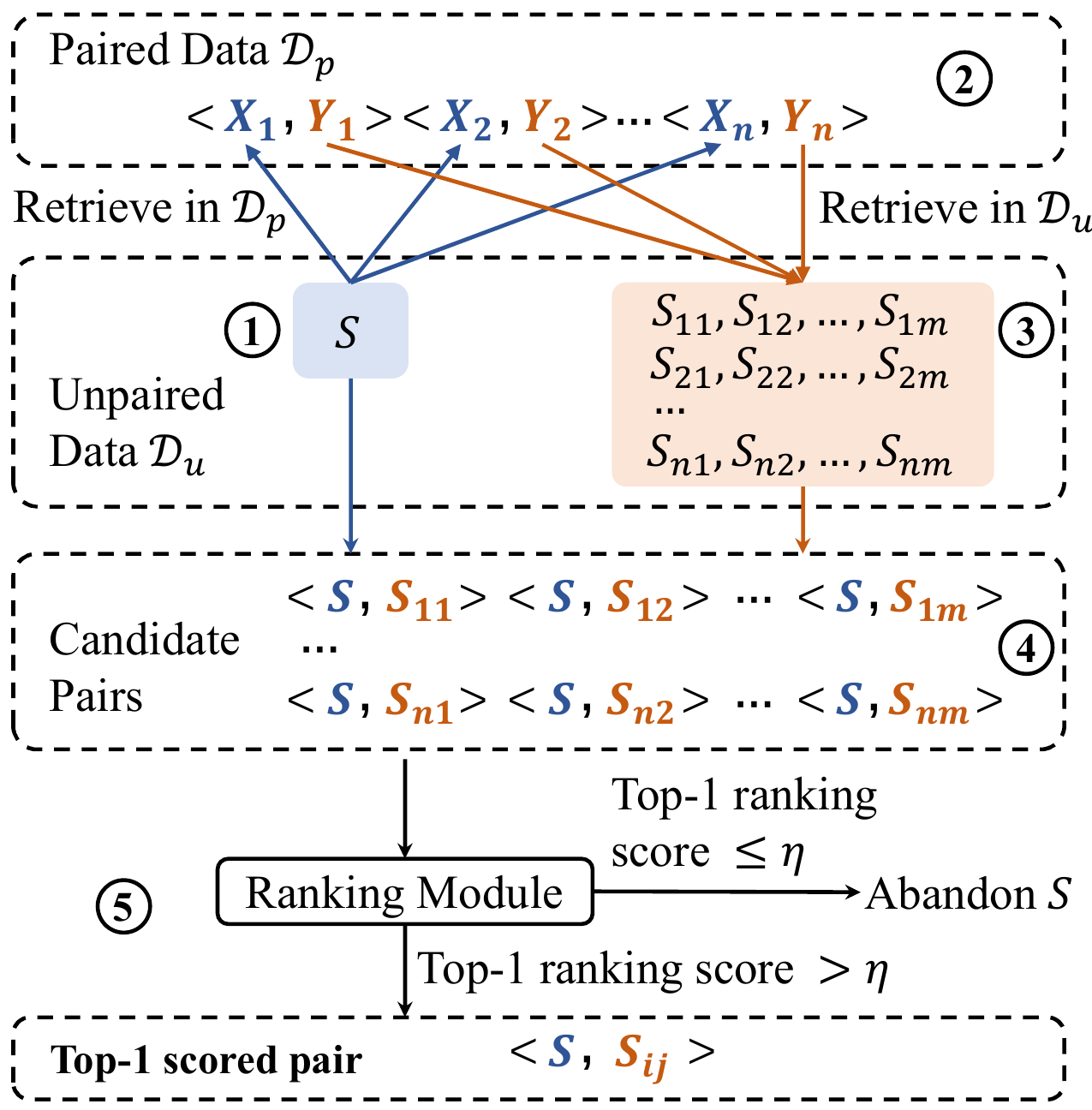}
  \caption{Framework of data-level distillation. (1) The sentence $S$ is randomly selected in the unpaired data $\mathcal{D}_u$. (2) A set of posts $X_1, \dots, X_n$ that are similar to $S$ are retrieved from the paired data $\mathcal{D}_p$. (3) Each corresponding response $Y_i$ is then used to retrieve $m$ sentences $S_{i1}, \dots, S_{im}$ that are similar to $Y_i$ from $\mathcal{D}_u$. (4) Then $n \times m$ candidate pairs can be formed by grouping $S$ with each sentence: $\langle S, S_{ij} \rangle$, ($i=1,\dots,n$, $j=1,\dots,m$). (5) A ranking module is used to rank these candidate pairs.}
  \label{fig:model_arch}
\end{figure}

The data-level distillation in our method aims at constructing a set of new post-response pairs $\mathcal{D}_a$ by matching non-parallel sentences retrieved from unpaired data $\mathcal{D}_u$. Specifically, $\mathcal{D}_p$ consists of $N$ post-response pairs: $\mathcal{D}_p=\{\langle X_i,Y_i \rangle\}_{i=1}^N$, in which $X_i$ and $Y_i$ is the post and response, respectively, and $\mathcal{D}_u$ consists of $M$ non-parallel sentences: $\mathcal{D}_u=\{S_i\}_{i=1}^M$. Note that $M$ is usually much larger than $N$ because non-parallel sentences are generally easier to collect.

Further, the output of our data-level distillation process is a set of augmented post-response pairs: $\mathcal{D}_a=\{\langle X'_i, Y'_i \rangle \}_{i=1}^{K}$, in which both the post and response come from the unpaired dataset $\mathcal{D}_u$, i.e., $X'_i \in \mathcal{D}_u$ and $Y'_i \in \mathcal{D}_u$ for $i=1, \dots, K$.

The data-level distillation involves two major processes: 1) constructing candidate pairs and 2) filtering low-quality candidates. The whole framework is shown in Figure~\ref{fig:model_arch} and detailed below.

\subsection{Constructing Candidate Pairs}\label{sec:construct_fp}

We first construct candidate dialogue pairs with the help of some post-response pairs $\langle X_i, Y_i \rangle$ selected from $\mathcal{D}_p$. The basic intuition is that sentences that are similar to post $X_i$ can usually be responded with sentences that are similar to the corresponding response $Y_i$. Candidate dialogue pairs can be then constructed by anchoring sentences in $\mathcal{D}_u$ using $\langle X_i, Y_i \rangle$.

The construction of candidate pairs starts by randomly selecting a sentence $S$ from the unpaired dataset $\mathcal{D}_u$. We then treat $S$ as a candidate post, and it is used to retrieve $n$ posts $X_i$ ($1 \leq i \leq n$) that are similar to $S$ from the paired data $\mathcal{D}_p$. In this study, the sentence retrieval process is implemented based on the Okapi BM25 algorithm, which scores the similarity of input sentences using bag-of-words features. Then the corresponding $n$ post-response pairs $\langle X_i, Y_i \rangle$ ($1 \leq i \leq n$) are extracted from $\mathcal{D}_p$. For each response $Y_i$, we further retrieve $m$ sentences $S_{ij}$ ($1 \leq j \leq m$) that are similar to $Y_i$ from the unpaired dataset $\mathcal{D}_u$. These sentences $S_{ij}$ can then serve as candidate responses to the original sentence $S$, and therefore $n \times m$ candidate pairs $\langle S, S_{ij} \rangle$, ($ 1 \leq i \leq n$, $ 1 \leq j \leq m$) are generated. Moreover, for each candidate pair $\langle S, S_{ij} \rangle$, we name the post-response pair $\langle X_i, Y_i \rangle$ in $\mathcal{D}_p$ that are used to produce $\langle S, S_{ij} \rangle$ as its \emph{``anchor pair''} since it anchors the sentences $S$ and $S_{ij}$ from $\mathcal{D}_u$. 

Note that we have explored other variants of the above process, such as treating the initial sentence $S$ as a candidate response rather than a candidate post or utilizing more advanced text retrieval methods to extract similar sentences. However, we notice little difference in neither the quality of the final augmented pairs nor the performance improvement brought to the dialogue models. 

\subsection{Filtering Candidate Pairs}\label{sec:cand_filter}
In order to enhance the quality of the augmented data, we propose to filter out low-quality pairs using a ranking module, which calculates a score for each candidate pair obtained above. Specifically, high-quality pairs that are fluent and coherent are expected to receive high scores. In this study, we implement the score function as a text matching model, which is built by fine-tuning a pre-trained BERT model on the paired dataset $\mathcal{D}_p$. Negative samples are constructed by replacing the original responses using randomly sampled sentences from $\mathcal{D}_p$. The ranking score for each input pair is calculated as the matching score produced by the matching model.

In this study, we follow a quite rigorous policy to select the final augmented pairs in $\mathcal{D}_a$. For each sample sentence $S$ from $\mathcal{D}_u$, we only extract the top-1 scored pair $\langle S, S_{ij} \rangle$ among all its $n \times m$ candidate pairs, and $\langle S, S_{ij} \rangle$ is added to $\mathcal{D}_a$ only when its matching score exceeds a certain threshold $\eta (0.9 \leq \eta)$. We repeat the above procedures with newly sampled sentences from $\mathcal{D}_u$ until a desired number of augmented pairs in $\mathcal{D}_a$ are obtained. The whole data-level distillation process in our method is summarized in Algorithm~\ref{alg:process}.

Note that the matching model used in the ranking process can also be directly used to align sentences from the unpaired dataset $\mathcal{D}_u$. Specifically, for a sampled sentence $S$ from $\mathcal{D}_u$, we can treat all other sentences in $\mathcal{D}_u$ as its candidate response and select an augmented pair by ranking all these candidates. Although theoretically possible, this approach is practically infeasible considering the large amount of sentences in $\mathcal{D}_u$ and the tremendous computational load to rank these candidates. Note that previous works on effective ranking (such as \citet{henderson2017efficient,henderson2020convert}) can not be directly adapted to this study because our ranking model does not use dot-product scoring function.

\begin{algorithm}[tb]
\small
\caption{Data-level distillation process} \label{alg:process}
\hangindent 3em \textbf{Input}:
A set of unpaired data $\mathcal{D}_u$=$\{S_i\}_{i=1}^M$,
a set of paired data $\mathcal{D}_p$=$\{\langle X_i,Y_i \rangle \}_{i=1}^N$, a threshold $\eta$.

\textbf{Output}:
Augmented dialogue pairs $\mathcal{D}_a$=$\{\langle X'_i, Y'_i \rangle\}_{i=1}^{K}$.

\begin{algorithmic}[1] 
\STATE $\mathcal{D}_a$ $\leftarrow$ Empty set
\WHILE{$|\mathcal{D}_a| < K$}
\STATE $\widetilde{\mathcal{D}}_a$ $\leftarrow$ Empty set
\STATE Sample a sentence $S \sim \mathcal{D}_u$.
\STATE Retrieve $n$ posts $\{X_i\}_{i=1}^{n}$ that are similar to $S$ in $\mathcal{D}_p$.
\STATE Get the responses $\{Y_i\}_{i=1}^{n}$ for $\{X_i\}_{i=1}^{n}$ from $\mathcal{D}_p$.
\FOR{each $Y_i$}
\STATE Retrieve $m$ sentences $\{S_{ij}\}_{j=1}^{m}$ that are similar to $Y_i$ in $\mathcal{D}_u$.
\STATE $\widetilde{\mathcal{D}}_a$ $\leftarrow$ $\widetilde{\mathcal{D}}_a \bigcup \{\langle S, S_{ij}\rangle\}_{j=1}^m$
\ENDFOR
\STATE Calculate the ranking score for each pair in $\widetilde{\mathcal{D}}_a$.
\STATE Extract the top-1 scored pair $\langle S, S_{ij} \rangle$ from $\widetilde{\mathcal{D}}_a$.
\IF{The ranking score of $\langle S, S_{ij} \rangle$ exceeds $\eta$}
\STATE $\mathcal{D}_a$ $\leftarrow$ $\mathcal{D}_a \bigcup \{\langle S, S_{ij}\rangle\}$
\ENDIF
\ENDWHILE
\end{algorithmic}
\end{algorithm}

\section{Model-level Distillation}\label{sec:train_with_teacher}

A straightforward way to improve a dialogue model with the augmented dialogue data is to directly merge the original paired data $\mathcal{D}_p$ with $\mathcal{D}_a$. However, this naive approach may lead to sub-optimal performance since the augmented pairs in $\mathcal{D}_a$ might not be as high-quality as these human-crafted pairs in $\mathcal{D}_p$. In this study, we apply the model-level distillation in the training process to prevent the dialogue models from being affected by the noise in $\mathcal{D}_a$. This approach can be used in both retrieval-based and generation-based dialogue models.

\subsection{Retrieval-based Dialogue Model}\label{sec:retrieval_model}

A retrieval-based dialogue model produces responses by retrieving a best matching sentence from the pre-collected dialogue dataset. Its key component is a matching function $\mathcal{P}_{\theta}(l|X, Y)$ that predicts whether a response $Y$ matches a given post $X$. Specifically, $l \in \{0,1\}$ is a matching label, where $l=1$ means $Y$ is a proper response for $X$ and $l=0$ otherwise. The model parameters $\theta$ can be learned by optimizing a negative log likelihood (NLL) loss defined as
\begin{equation}
    \begin{split}
    \mathcal{L}_{m-nll}(\theta)=&-(1-l) {\rm log}\mathcal{P}_{\theta}(0|X,Y) \\
    &- l {\rm log} \mathcal{P}_{\theta}(1|X,Y)
    \end{split}
    \label{eq:match_nll}
\end{equation}

In this study, we formalize the matching function using the BERT model~\cite{devlin2018bert,Whang2020Effective}. A teacher model $\mathcal{P}_{\theta_t}(l|X, Y)$ is first obtained by optimizing the NLL loss $\mathcal{L}_{m-nll}(\theta_t)$ on the paired dataset $\mathcal{D}_p$. After the training is completed, the teacher model is fixed and used to compute a knowledge distillation (KD) loss~\cite{kim2016sequence} as
\begin{equation}
    \mathcal{L}_{m-kd}(\theta)= - \sum_{i=0}^1 \mathcal{P}_{\theta_t}(i|X,Y) \cdot {\rm log}\mathcal{P}_{\theta}(i|X,Y).
    \label{eq:matching_kd_loss}
\end{equation}
The final matching model is trained on the following loss:
\begin{equation}
    \mathcal{L}_{M}(\theta)=\mathcal{L}_{m-nll}(\theta) + \alpha_m \mathcal{L}_{m-kd}(\theta),
    \label{eq:matching_total_loss}
\end{equation}
where the loss $\mathcal{L}_{M}(\theta)$ is evaluated using $\mathcal{D}_p \bigcup \mathcal{D}_a$  and $\alpha_m$ is used to balance these two losses. 
\subsection{Generation-based Dialogue Model}\label{sec:generation_model}

A generation-based dialogue model tries to capture the distribution of the response sentences $Y$ given the post sentence $X$, i.e., $\mathcal{P}_\phi(Y|X)$, which can be formalized as
\begin{equation}
    \mathcal{P}_{\phi}(Y|X)=\prod_{i=1}^{|Y|} P_{\phi}(y_i|y_{<i},X),
\end{equation}
where $|Y|$ is the length of $Y$, $y_{<i}=y_1\cdots y_{i-1}$ is the token sequence before $y_i$. The model parameters $\phi$ can be learned by optimizing the NLL loss:
\begin{equation}
    \mathcal{L}_{g-nll}(\phi)=- \sum_{i=1}^{|Y|} {\rm log}P_{\phi}(y_i|y_{<i},X).
    \label{eq:g_nll}
\end{equation}

In this study, we parameterize the dialogue generation model using the Transformer-based encoder-decoder framework~\cite{vaswani2017attention,golovanov2019large,zheng2019pre}. Similar to the retrieval-based approach, a teacher model is first obtained by optimizing the NLL loss $\mathcal{L}_{g-nll}$ on the paired dataset $\mathcal{D}_p$ and the trained teacher model is used to compute a KD loss as
\begin{equation}
    \begin{split}
    \mathcal{L}_{g-kd}(\phi)=- &\sum_{i=1}^{|Y|} \sum_{j=1}^{|\mathcal{V}|} P_{\phi_t}(y_i=j|y_{<i},X) \\
    &\times {\rm log}P_{\phi}(y_i=j|y_{<i},X),
    \end{split}
    \label{eq:g_kd}
\end{equation}
where $|\mathcal{V}|$ denotes the size of the vocabulary and $\phi_t$ is the parameter of the teacher model, which is fixed.

The final loss for the generation model is:
\begin{equation}
    \mathcal{L}_{G}(\phi)=\mathcal{L}_{g-nll}(\phi) + \alpha_g \mathcal{L}_{g-kd}(\phi),
    \label{eq:generation_total_loss}
\end{equation}
where the loss $\mathcal{L}_{G}(\theta)$ is evaluated using $\mathcal{D}_p \bigcup \mathcal{D}_a$ and $\alpha_g$ is used to balance these two losses. 

\section{Experiment}\label{sec:exp}
\subsection{Dataset}
The evaluation of our method is performed on a corpus collected from Weibo\footnote{https://www.weibo.com}. Specifically, the paired data $\mathcal{D}_p$ contains 300K post-response pairs, which are made up of Weibo posts and their following replies. All these pairs are manually filtered with annotators by removing ungrammatical sentences and incoherent dialogues. The unpaired data $\mathcal{D}_u$ contains about 2 million posts on Weibo that do not have replies. Non-fluent sentences in $\mathcal{D}_u$ are filtered out using a set of heuristic rules. Further, two additional sets of paired data are also prepared to validate and test the dialogue models, with 10K and 5K pairs respectively. These dialogue pairs are collected and manually filtered using the same criterion as $\mathcal{D}_p$. 

\subsection{Implementation Details}

\textbf{Data-level Distillation:}
We implement the retrieval module in Section~\ref{sec:construct_fp} using the Lucene library \footnote{https://lucene.apache.org/core/}, and set the value of both $n$ and $m$ to 5. The matching model used in Section~\ref{sec:cand_filter} is fine-tuned with $\mathcal{D}_p$ for three epochs based on the pretrained BERT-base model~\cite{devlin2018bert}. The hyper-parameter setting of the matching model follows the work of \citet{devlin2018bert}.

\textbf{Model-level Distillation:}
For the \emph{retrieval-based dialogue model}, the matching model used in Section~\ref{sec:cand_filter} is directly used as the teacher model to calculate the KD loss (Eq.~\ref{eq:matching_kd_loss}). The final retrieval-based dialogue model is initialized with the pre-trained BERT-base weights and fine-tuned using the loss in Eq.~\ref{eq:matching_total_loss} for 2 epochs on $\mathcal{D}_p \bigcup \mathcal{D}_a$. The value of $\alpha_m$ in Eq.~\ref{eq:matching_total_loss} is set to 1.

For the \emph{generation-based dialogue model}, the encoder and decoder share the same set of parameters, which is initialized using a pretrained GPT model~\cite{wang2020chinese}. The teacher model uses the same architecture and it is fine-tuned using the paired dataset $\mathcal{D}_p$ for 15 epochs on the NLL loss (Eq.~\ref{eq:g_nll}). The final generative dialogue model is first initialized using the pre-trained GPT weights and fine-tuned using the loss in Eq.~\ref{eq:generation_total_loss} for 50 epochs on $\mathcal{D}_p$ and  $\mathcal{D}_a$. The value of $\alpha_g$ in Eq.~\ref{eq:generation_total_loss} is set to 1. Moreover, the GPT model used in the initialization phase is trained on a corpus collected from various Chinese novels. This corpus contains about 0.5 billion tokens and a character-level vocabulary of size 13,084.

See Appendix \ref{sec:appendix} for more details of the model setting and reproduction guidance. The data and code for all experiments can be downloaded from the link\footnote{\url{https://github.com/njuzrs/dialogue\_distillation}}.

\begin{table*}[ht]
\setlength\tabcolsep{2.1pt} 
\centering
\setlength\tabcolsep{6pt}
\setlength{\belowcaptionskip}{-0.2cm}
\begin{tabular}{p{60pt}p{26pt}p{26pt}p{26pt}p{26pt}|p{26pt}p{26pt}p{26pt}p{26pt}|p{26pt}p{26pt}}
\toprule
Model          & \multicolumn{4}{c}{Distinct-1,2,3,4} & \multicolumn{4}{c}{Novelty-1,2,3,4} & Flu. & Coh. \\
\midrule            
CVAE & $0.178^\ddag$ & $09.40^\ddag$ & $34.54^\ddag$ & $60.73^\ddag$ &$00.25^\ddag$ & $08.47^\ddag$ & $25.45^\ddag$ & $40.62^\ddag$ & $1.529^\ddag$ & $0.862^\ddag$ \\
BT   & $0.193^\ddag$ & $12.42^\ddag$ & $43.43^\ddag$ & $70.38^\ddag$ &$03.07^\ddag$ & $21.66^\ddag$ & $35.28^\ddag$ & $45.18^\ddag$ & $1.771^\ddag$ & $\underline{1.408}^\dag$ \\
SP   & \textbf{0.228}& $11.56^\ddag$ & $37.76^\ddag$ & $57.73^\ddag$ &$18.48^\ddag$ & $46.65^\ddag$ & $73.56^\ddag$ & $87.79^\ddag$ & $1.839^\ddag$ & $0.777^\ddag$ \\
\midrule
DL $\eta$=0.90 & $\underline{0.226}^\ddag$ & \textbf{13.72} & \textbf{48.24} & \textbf{76.21} & \textbf{23.76} & \textbf{55.95} & \textbf{80.64} & \textbf{92.10} & $1.835^\ddag$ & $1.183^\ddag$ \\
DL $\eta$=0.95 & $0.224^\ddag$ & $\underline{13.44}^\ddag$ & $\underline{47.51}^\ddag$ & $\underline{75.55}^\ddag$ & $\underline{22.81}^\ddag$ & $\underline{55.51}^\ddag$ & $\underline{80.37}^\ddag$ & $\underline{91.97}^\ddag$ & $\underline{1.856}^\dag$ & $1.358^\ddag$ \\
DL $\eta$=0.99 & $0.213^\ddag$ & $12.61^\ddag$ & $45.06^\ddag$ & $72.87^\ddag$ & $21.59^\ddag$ & $54.40^\ddag$ & $79.69^\ddag$ & $91.62^\ddag$ & \textbf{1.877} & \textbf{1.428} \\
\midrule
$\mathcal{D}_p$(human) & 0.199& 13.51 & 47.70 & 75.52 & \multicolumn{4}{c|}{N/A} & 1.868 & 1.617 \\
\bottomrule
\end{tabular}
\caption{Automatic and manual evaluation on the quality of augmented pairs produced by different methods. The bottom row corresponds to the human filtered dialogue pairs in $\mathcal{D}_p$. The best results are in \textbf{bold}, and the second best results are \underline{underlined} (except ``human''). Significant tests between the best model and others were performed using t-test. $\dag$ and $\ddag$ indicates $p$-value \textless~0.01 and 0.001, respectively.}
\label{tab:eval_augmented}
\end{table*}

\subsection{Evaluating Augmented Dialogue Pairs}

\subsubsection{Baselines}\label{sec:augment_baseline}
We first evaluate the quality of the augmented pairs generated by our \underline{D}ata-\underline{L}evel (\textbf{DL}) distillation process. Three different matching thresholds $\eta$ in Algorithm~\ref{alg:process} are tested, i.e., $\eta$ = 0.90, 0.95, 0.99. Several strong baselines are also compared:

\textbf{CVAE}: A \underline{CVAE}-based model as proposed by \citet{li2019insufficient} is trained on the paired data $\mathcal{D}_p$. Augmented pairs are generated by sampling different latent codes.

\textbf{BT}: Augmented pairs are generated by \underline{B}ack \underline{T}ranslating (i.e., translate Chinese to English and then translate back to Chinese) the post sentences of the dialogue pairs in $\mathcal{D}_p$. The translation is done via the Google Translate API.

\textbf{SP}: A variant of our method is implemented by first \underline{S}ampling a post-response \underline{P}air $\langle X,Y \rangle$ from $\mathcal{D}_p$, and then retrieving a best-matching post and response from the unpaired data $\mathcal{D}_u$ using $X$ and $Y$ as the query, respectively. An augmented pair is constructed by pairing the retrieved post and response sentence without the ranking process. 

Note that there are two major differences between the baseline SP and our data-level distillation process: 1) the baseline SP starts with a dialogue pair $\langle X,Y \rangle$ sampled from $\mathcal{D}_p$ rather than a candidate post sampled from $\mathcal{D}_u$; 2) The ranking process is not used in the baseline SP to further filter the candidate pairs.

\subsubsection{Metrics}\label{sec:ap_human_metric}
The automatic evaluation of augmented dialogue pairs uses the following metrics: 1) \emph{\textbf{Distinct}}~\cite{li2015diversity} is used to measure the proportion of unique n-grams in the augmented dialogue pairs ($n$=1,2,3,4); 2) \emph{\textbf{Novelty}}~\cite{wang2018sentigan} is used to measure the proportion of new n-grams in the augmented dialogue pairs ($n$=1,2,3,4), i.e., n-grams that are covered by the augmented dialogue pairs but are not shown in the paired dataset $\mathcal{D}_p$. A higher novelty score means the augmented dialogue pairs contain more ``novel'' contents.

Manual evaluation is also used to evaluate the quality of augmented dialogue pairs. Three annotators are employed to rate these pairs from two aspects: 1) \emph{Fluency} (\textbf{Flu.}): whether the augmented pairs are fluent; 2) \emph{Coherency} (\textbf{Coh.}): whether the response is coherent with the post so that they make a plausible dialogue pair. The rating scale for each measure is of (0, 1, 2), in which 0 means worst and 2 best.

\subsubsection{Results}
Each data augmentation method introduced above are used to generate 300K augmented dialogue pairs, and on which automatic evaluation is performed. Further, manual evaluation is carried out on 200 dialogue pairs that are randomly sampled from these augmented data, and the inter-rater agreement between annotators is measured using the Fleiss's kappa $\kappa$~\cite{randolph2005free}. The $\kappa$ value for \emph{Fluency} and \emph{Coherency} is 0.69 (substantial agreement), and 0.42 (moderate agreement), respectively. Note that this evaluation is purely regarding the augmented dialogue data, without considering any dialogue model training.

The evaluation results in Table~\ref{tab:eval_augmented} demonstrate that the augmented dialogue data produced by our method outperform all the baselines in almost all the metrics. We can further observe that: \textbf{1})  Our method obtains similar scores on all the metrics compared to these human-produced and filtered dialogue pairs in $\mathcal{D}_p$. This indicates that the augmented dialogue pairs generated by our method are of high quality. We present some examples of the augmented pairs together with their associated anchor pairs in Table~\ref{tab:example_pair}. \textbf{2}) The matching threshold $\eta$ can be used to trade off between the coherency and diversity of the augmented dialogue pairs. Specifically, a higher $\eta$ value improves Fluency and Coherency scores but hurts Distinct and Novelty scores of the augmented pairs.

\begin{table*}[ht]
\small
\centering
\setlength\tabcolsep{2pt}
\begin{tabular}{lll}  
\toprule
 & \multicolumn{1}{c}{Augmented pairs} & \multicolumn{1}{c}{Associated anchor pairs from $\mathcal{D}_p$}\\
\midrule
{Post} & I'm almost moved to cry  \small{(我已经快感动地哭了)} & I am so moved today! \small{(今天感动得快哭了！)}\\
{Resp} & What happened there? \small{(发生什么事情呢？)} & What happen \small{(发生什么事)} \\
\midrule
Post & I like it, men should be like this \small{(这话我喜欢。男人就该这样)} & I like this types of man \small{(喜欢这样的男人)} \\
Resp & I like it too, just as you do \small{(我也喜欢。跟你一样)} & Your taste is just like mine \small{(怎么跟我喜欢的一样)} \\
\midrule
Post & I liked to play it when I was young \small{(小时候很喜欢玩)} & My favorite toy in kindergarten \small{(幼儿园最喜欢玩的)}\\
Resp & I have also played, it's so cute \small{(表示有幸玩过，很萌哒)} & I have also played, lol \small{(我也玩过哒)} \\
\bottomrule
\end{tabular}
\caption{Example pairs produced by the proposed data augmentation method. The associated anchor pairs are also shown. More examples are shown in Appendix \ref{appendix:examples}}
\label{tab:example_pair}
\end{table*}

\subsection{Evaluating Dialogue Models}

\subsubsection{Baselines}
We evaluate the benefit of the augmented dialogue data in both retrieval-based and generation-based dialogue models. Specifically, 300K augmented dialogue pairs are generated using these three baselines introduced in Section~\ref{sec:augment_baseline}, and the model-level distillation process as introduced in Section~\ref{sec:train_with_teacher} is used to train the dialogue models. We denote these three dialogue model baselines as \textbf{CVAE+ML}, \textbf{BT+ML}, and \textbf{SP+ML}, respectively. Note that the notation ``ML'' means that the \underline{M}odel-\underline{L}evel distillation is used. Moreover, besides comparing to different data augmented methods as introduced in Section~\ref{sec:augment_baseline}, several other competitive dialogue model baselines are also tested:

\textbf{Teacher}: Training the dialogue models on the paired data $\mathcal{D}_p$ with the NLL loss. Note that this setting produces the teacher models used in Section~\ref{sec:train_with_teacher}.

\textbf{AP}: Training dialogue models only on the \underline{A}ugmented \underline{P}airs $\mathcal{D}_a$ with the NLL loss.

\textbf{UP+PreT}: First fine-tuning the pre-trained GPT (with the NLL loss in Eq.~\ref{eq:g_nll}) or BERT-base model (with the MLM loss~\cite{devlin2018bert}) on the \underline{U}n\underline{P}aired Data $\mathcal{D}_u$, and then using these fine-tuned weights to initialize the dialogue models, which are further fine-tuned on $\mathcal{D}_p$ with the NLL loss.

\textbf{NP+ML}: Sampling 300K pairs from a set of Weibo dialogues that are not manually filtered and use these ``\underline{N}oisy \underline{P}airs'' as the augmented pairs. The model-level distillation process introduced in Section~\ref{sec:train_with_teacher} is used to train this baseline.

We denote our method as \textbf{DL+ML} since it trains the dialogue model using both the data-level and model-level distillation. The threshold $\eta$ in Algorithm~\ref{alg:process} is set to 0.95 for a better trade-off between the coherency and diversity of the augmented data. Further, we also test another method to work with data-level distillation (i.e., utilizing $\mathcal{D}_a \bigcup \mathcal{D}_p$): \textbf{DL+PreT}, i.e., first pre-train the dialogue model on $\mathcal{D}_a$ and then fine-tune on $\mathcal{D}_p$ with the NLL loss.

Further, we also performed several ablation tests on our method to validate the effect of each component:
1) training dialogue models on $\mathcal{D}_p \bigcup \mathcal{D}_a$ using only the NLL loss, i.e., without the model-level distillation (\textbf{w/o ML});
2) training dialogue models only on the paired data $\mathcal{D}_p$ using $\mathcal{L}_M(\theta)$ or $\mathcal{L}_G(\phi)$, i.e., the data-level distillation are not used (\textbf{w/o DL});
3) training dialogue models on the augmented data $\mathcal{D}_a$ using $\mathcal{L}_M(\theta)$ or $\mathcal{L}_G(\phi)$, i.e., the paired data $\mathcal{D}_p$ are not used (\textbf{w/o PD});
4) generating $\mathcal{D}_a$ without the ranking module (\textbf{w/o Ranking}), i.e., the candidate pairs are used as the augmented data without filtering.

Note that all the baselines and ablation models are initialized with pre-trained GPT or BERT-base weights.

\subsubsection{Metrics}
The retrieval-based dialogue models are evaluated using the following metrics:
1) \emph{Mean Average Precision} (\textbf{MAP}): the average rank of the reference responses;
2) \emph{$\bm{R_{10}@k}$}: the recall of the reference response being in the top-$k$ ranked candidates ($k$=1,2,5) when given 10 candidates in total.

The generation-based dialogue models are evaluated both automatically and manually. Specifically, the following automatic metrics are used:
1) \emph{Perplexity} (\textbf{PPL}) which measures how the model fits the test data;
2) \emph{\textbf{BLEU}} which evaluates the overlap of n-grams (n=1,2) between the generated and reference responses;
3) \emph{Distinct} (\textbf{Dist.}) measures the proportion of unique n-grams in the generated responses (n=1,2).
Manual evaluation is also performed for the generated dialogue responses following the same protocol as introduced in Section~\ref{sec:ap_human_metric}.

\begin{table}[!tbh]
\centering
\setlength\tabcolsep{2pt}
\begin{tabular}{lccccc}  
\toprule
Model    & MAP  & $R_{10}@1$ & $R_{10}@2$ & $R_{10}@5$  \\
\midrule    
Teacher       & 80.2 & 69.7 & 82.1 & 95.1 \\
AP       & 76.5 & 65.1 & 78.0 & 92.1 \\
UP+PreT    & 80.6 & 70.3 & 82.6 & \textbf{95.3} \\  
NP+ML    & 80.8 & 70.5 & 82.9 & 95.2 \\     
CVAE+ML     & 80.3 & 69.8 & 82.5 & 94.9 \\
BT+ML       & 80.3 & 69.8 & 82.0 & 95.2 \\     
SP+ML       & 80.4 & 70.0 & 82.0 & 95.2 \\ 
\midrule
DL+PreT    & 80.7 & 70.2 & 82.7 & \textbf{95.3} \\
DL+ML  & \textbf{81.0} & \textbf{70.8} & \textbf{83.1} & \textbf{95.3} \\ 
\midrule
w/o ML   & 80.4 & 69.9 & 82.5 & 95.0 \\
w/o DL   & 80.5 & 70.1 & 82.3 & 95.1 \\
w/o PD   & 79.5 & 68.9 & 81.3 & 94.1 \\
w/o Ranking  & 80.5 & 70.1 & 82.5 & 95.2 \\
\bottomrule
\end{tabular}
\caption{Automatic evaluation for retrieval-based dialogue models with different training and data augmentation methods.}
\label{tab:retrival_res}
\end{table}

\begin{table}[!tbh]
\setlength{\belowcaptionskip}{-0.3cm}
\setlength\tabcolsep{2.1pt} 
\centering
\setlength\tabcolsep{2pt}
\begin{tabular}{p{62pt}p{25pt}p{30pt}p{25pt}p{25pt}p{30pt}}
\toprule
Model    & PPL   & \multicolumn{2}{c}{BLEU-1,2}  & \multicolumn{2}{c}{Dist.-1,2}  \\
\midrule 
Teacher             & $23.9^\ddag$ & $12.25^\ddag$&$6.61^\ddag$  & $3.83^\ddag$&$29.69^\ddag$ \\
AP             & $50.0^\ddag$ & $10.86^\ddag$&$5.52^\ddag$  & $3.29^\ddag$&$23.37^\ddag$ \\
UP+PreT          & $24.0^\ddag$ & 12.60&$6.81^\dag$  & $3.99^\ddag$&$30.50^\ddag$  \\  
NP+ML     & $23.1^\ddag$ & $11.63^\ddag$&$6.25^\ddag$  & $3.99^\ddag$&$28.47^\ddag$  \\     
CVAE+ML   & $23.9^\ddag$ & $12.27^\ddag$&$6.59^\ddag$  & $3.73^\ddag$&$26.75^\ddag$  \\
BT+ML     & $23.8^\ddag$ & $11.93^\ddag$&$6.48^\ddag$  & $3.84^\ddag$&$27.38^\ddag$  \\     
SP+ML     & $23.6^\ddag$ & $12.47^\ddag$&$6.74^\ddag$  & 4.04&$30.66^\ddag$  \\ 
\midrule
DL+PreT & $23.7^\ddag$ & \textbf{12.66}  &6.92  & $3.95^\ddag$&$30.30^\ddag$ \\
DL+ML  &  \textbf{22.6} & $12.42^\ddag$&\textbf{6.93}  &  \textbf{4.13}&\textbf{31.39} \\ 
\midrule
w/o ML    & $23.3^\ddag$ & $12.30^\ddag$&$6.65^\ddag$  & 4.06&$30.89^\ddag$ \\
w/o DL         & $23.5^\ddag$ & $12.54^\dag$ &6.88  & $3.96^\ddag$&$29.79^\ddag$ \\
w/o PD         & $26.7^\ddag$ & $11.08^\ddag$&$5.86^\ddag$  & $3.48^\ddag$&$26.84^\ddag$ \\
w/o Ranking    & $22.8^\ddag$ & $12.54^\ddag$&$6.78^\ddag$  & $3.90^\ddag$&$28.93^\ddag$ \\
\bottomrule
\end{tabular}
\caption{Automatic evaluation results for generation-based dialogue models with different training and data augmentation methods. Significance tests between the best model and others were performed using t-test with booststrap resampling \cite{koehn-2004-statistical}. $\dag$ and $\ddag$ indicates $p$-value \textless~0.005 and 0.001, respectively.}
\label{tab:generation_res}
\end{table}

\subsubsection{Results}
Automatic evaluation for each dialogue model is performed on 5K test data (see Table~\ref{tab:retrival_res} and Table~\ref{tab:generation_res} for the results), and manual evaluation is performed using 200 pairs that are randomly sampled from these test data (see Table~\ref{tab:gen_human_res} for the results). The $\kappa$ value for the \emph{Fluency} and \emph{Coherency} annotation is 0.9 (substantial agreement) and 0.56 (moderate agreement), respectively.

Our method outperforms all the baselines in almost all the metrics for both retrieval-based and generation-based dialogue models. We can further observe that:
\textbf{1}) The dialogue models that utilize unpaired data $\mathcal{D}_u$ (e.g. DL+ML, DL+PreT, UP+PreT) generally outperform the models that are only trained on $\mathcal{D}_p$ (e.g., Teacher, CVAE+ML). This demonstrates that utilizing unpaired data is more effective at improving the performance of dialogue models;
\textbf{2}) Training the dialogue models on the merged data $\mathcal{D}_p \bigcup \mathcal{D}_a$ without utilizing the model-level distillation (i.e., w/o ML) brings little or no performance improvements compared to directly training on $\mathcal{D}_p$ (i.e., Teacher). This verifies the effectiveness of the model-level distillation process proposed in our method;
\textbf{3}) When the model-level distillation is employed, the augmented data produced by our data-level distillation process (i.e., DL+ML) can better improve the performance of dialogue models compared to the augmented data produced by other data augmentation methods (e.g. CVAE+ML, NP+ML, SP+ML, BT+ML). This verifies the effectiveness of the data-level distillation process proposed in our study.

\begin{table}[!tbh]
\setlength{\abovecaptionskip}{0.25cm}
\setlength{\belowcaptionskip}{-0.3cm}
\centering
\begin{tabular}{llllll}  
\toprule
Model    &  Flu.  & Coh.   \\
\midrule 
Teacher       & $1.968^\ddag$ & $1.432^\ddag$ \\
AP       & 1.985 & $1.417^\ddag$ \\
UP+PreT    & $1.957^\ddag$ & 1.500 \\  
NP+ML       & $1.967^\ddag$ & $1.473^\dag$ \\     
CVAE+ML     & $1.977^\dag$ & $1.475^\dag$ \\
BT+ML       & $1.957^\ddag$ & 1.503 \\     
SP+ML       & $1.973^\dag$ & $1.453^\ddag$ \\ 
\midrule
DL+PreT    & 1.975 & $1.492^\dag$ \\
DL+ML     & \textbf{1.993} & \textbf{1.518}\\ 
\bottomrule
\end{tabular}
\caption{Manual evaluation for generation-based dialogue models. Significant tests between the best model and others were performed using t-test. $\dag$ and $\ddag$ indicate $p$-value \textless~0.05 and 0.01, respectively.}
\label{tab:gen_human_res}
\end{table}

\section{Conclusion}
This paper presents a novel dialogue distillation method that consists of two processes, i.e., 1) a data augmentation process to construct new post-response pairs from unpaired data and 2) a model distillation process that distills a teacher model trained on the original data to the augmented data. Automatic and manual evaluation shows that our method can produce high-quality post-response pairs that are both coherent and content-rich, which can be further used to improve the performance of competitive baselines. Our method may inspire other research in low-resource NLP tasks. 

\section*{Acknowledgments}
This work was jointly supported by the NSFC projects (Key project with No. 61936010 and regular project with No. 61876096), and the Guoqiang Institute of Tsinghua University, with Grant No. 2019GQG1.
We thank THUNUS NExT Joint-Lab for the support.

\bibliography{emnlp2020}
\bibliographystyle{acl_natbib}

\appendix

\section{Implementation Details of Dialogue Models}
\label{sec:appendix}
\textbf{Retrieval-based dialogue model:} For the retrieval-based dialogue models, we implement the matching models by fine-tuning the BERT-base model \cite{devlin2018bert}, which contains 12 Transformer layers with 768-dimensional hidden states. The feed-forward layer's inner states are 3,072 dimensions, and the multi-head attention layer involves 12 attention heads. The vocabulary size is 21,128, and the max sequence length is set to 512. We use the Adam optimizer ($\beta_1=0.9$, $\beta_2=0.999$ and $\epsilon=10^{-8}$) with a learning rate of 2e-5, the batch size is set to 32 and the warm-up step is set to 2000. Moreover, We fine-tune both the teacher and student models for three epochs.

\textbf{Generation-based dialogue model:} For the generation-based dialogue models, we share the weights of the encoder and decoder in each dialogue model and initialize these weights using a pre-trained GPT model \cite{radford2018improving}. Specifically, the GPT model we used is pre-trained on a dataset collected from a set of Chinese novels that cover various genres (including Comedy, Romance, Mystery). The final pre-training corpus contains about 0.5 billion tokens. Moreover, we use the character-level vocabulary of size 13,084, and the context length is set to 512. Our model contains a total number of 191.01M parameters, and the pre-training process lasts for a week on 8 GTX1080Ti GPUs. 

When fine-tuning our dialogue models, the teacher model is trained for 15 epochs (about 12 hours), and the student model is trained for 50 epoch (about 40 hours) on 4 GTX1080Ti GPUs. Moreover, the batch size is set to 128, and the maximum learning rate is 6.25e-5. The training starts with a warm-up step of 1,000, and the learning rate is annealed proportionally to the inverse square root of the step number. The Adam optimizer is used with the parameter $\beta_1=0.9$, $\beta_2=0.98$ and $\epsilon=10^{-9}$. In the inference phase, we use the beam search with size 5. The length penalty is set to 1.6, and the maximum decoded sequence length is set to 50.

Note that because the pre-training approach is utilized in our model and baselines, we inherit most of the hyper-parameter settings from the previous studies of the pre-training model \citet{radford2018improving,devlin2018bert}, and skip the hyper-parameter tuning process. Moreover, for fair comparisons, we use a fixed set of the hyper-parameters in all our experiments (including all the ablation models and the Transformer-based baselines).

\section{More Augmented Dialogues Pairs}
\label{appendix:examples}
We provide more examples of the augmented pairs together with their associated anchor pairs in Table~\ref{tab:more_example_pair}.

\begin{table*}[htbp]
\small
\centering
\setlength\tabcolsep{1pt}
\begin{tabular}{ccc}  
\toprule
 & \multicolumn{1}{c}{Augmented pairs} & \multicolumn{1}{c}{Associated anchor pairs from $\mathcal{D}_p$}\\
\midrule
{post} & I am in Taiyuan, 24 years old, want to go to the Czech  & I am in Nanjing, 20 years old, want to go to the Czech \\
& (我在太原，24岁，想去捷克) & (我在南京，二十岁，想去捷克) \\
{resp} & I am in Henan, 22 years old, want to go to Lijiang & I am in Nanjing, 22 years old, want to go to Canada\\
& (我在河南，22岁，想去丽江) & (我在南京，22岁，想去加拿大) \\
\midrule
{post} & This love is strange and I can't understand. & I can't understand. \\
& (这相爱好奇怪，无法理解。) & (无法理解) \\
{resp} & It's not difficult to understand. They just need it. & Then don't understand. \\
& (不难理解，就是很需要。) & (那就不要理解) \\
\midrule
{post} & Completely denied the claim that clothes make the man ... & Clothes make the man \\
& (完全否定了人靠衣装这个说法···) & (人靠衣装马靠鞍啊) \\
{resp} & It's not true that clothes make the man! Man makes clothes! & Clothing is for beauties \\
& (人靠衣装这话是假的！是衣靠人装！) & (衣装毅然是给美女的) \\
\midrule
{post} & It seems wrong... The person I dreamed of do not miss me... & I think I will never find someone who treats me like you do\\
& (好像不对吧…我梦到的人不应该想我呀…) & (我想我应该再也找不到像你那样对我好的人了) \\
{resp} & As long as you know I miss you & As long as you know \\
& (你知道就好，想你了) & (你知道就好) \\
\midrule
{post} & Life is short, we should have fun. & Life is short and we should have fun \\
& (人生在世，需及时行乐。) & (人生在世需及时行乐) \\
{resp} & That makes sense, good morning! & That makes sense \\
& (说的挺有道理，早上好！) & (说的好像也挺好道理的) \\
\midrule
{post} & Men are really not easy. Sisters, be considerate! & To be honest, it's not easy. \\
& (男人们真心不容易啊。姐妹们体谅一下！) & (真心的不容易啊) \\
{resp} & It ’s not easy to do anything, is it? & Nothing is easy \\
& (做什么都不容易，不是么) & (什么都不容易呢) \\
\midrule
{post} & It is always difficult to make a choice & What is the most difficult problem? Choose it \\
& (人对于选择总是最难的) & (最难的难题是什么?选择吧) \\
{resp} & It is hard to give up your greed rather than worry & The most difficult problem is that you have to give up \\
& (难得不是放下烦恼而是放弃自己的贪念) & (最难得难题是属於自己却不得不放弃) \\
\midrule
{post} & Why are you always laughing so happily! & Why are you so happy \\
& (尼玛总是笑得那么开心干嘛！) & (干嘛心情这么开心) \\
{resp} & Laugh when you are happy. Laugh later when you are not. & I'll be unhappy later. I am enjoying my time \\
& (开心了就笑不开心了待会儿再笑。) & (待会儿就不开心了,抓紧时间) \\
\midrule
{post} & It's really cozy. I also want to go home & I really want to go home. Go back to my cozy island\\
& (真的好温馨。我也好想回家了) & (好想回家,回温暖的小岛) \\
{resp} & It's almost New Year, when are you on holiday? & When will you learn? Coming back for New Year \\
& (快过年了，你们什么时候放假呢？) & (要学习到什么时候呢?快回来过年啦) \\
\midrule
{post} & That's right. Work is the most annoying thing & Work is the most annoying thing\\
& (说的真对。上班什么的都最讨厌了) & (上班什么的最讨厌啦) \\
{resp} & I hate meetings. Meetings lead to overtime work! & Meeting is more annoying than work  \\
& (最讨厌开会，开会必加班！) & (比上班更讨厌的是开会) \\
\bottomrule
\end{tabular}
\caption{Augmented pairs produced by our data augmentation method. The associated anchor pairs are also given.}
\label{tab:more_example_pair}
\end{table*}

\end{CJK}

\end{document}